\documentclass{article}

\usepackage{arxiv}

\usepackage{amsmath}
\usepackage[backend=biber]{biblatex}   
\usepackage{nicefrac}
\usepackage{listings} 
\usepackage[usenames,dvipsnames]{xcolor}
\usepackage{graphicx}
\usepackage{multicol,lipsum}

\lstdefinelanguage{Julia}%
{%
morekeywords={abstract,break,case,catch,const,continue,do,else,elseif,%
		end,export,false,for,function,immutable,import,importall,if,in,%
		macro,module,otherwise,quote,return,switch,true,try,type,typealias,%
		using,while,
		fmiLoad,fmiSimulate,fmiPlot,fmiUnload,
		fmiInstantiate,fmiSetupExperiment,fmiEnterInitializationMode,fmiExitInitializationMode,fmiDoStep,fmiTerminate,fmiFreeInstance,
		sspLoad,sspSimulate,sspPlot,sspUnload,
		params, train, length,
		NeuralFMU,
		Chain, Dense	},%
	sensitive=true,%
	alsoother={$},%
	morecomment=[l]\#,%
	morecomment=[n]{\#=}{=\#},%
	morestring=[s]{"}{"},%
	morestring=[m]{'}{'},%
}[keywords,comments,strings]%

\usepackage[utf8]{inputenc} 
\usepackage[T1]{fontenc}   
\usepackage{hyperref}      
\usepackage{url}           
\usepackage{booktabs}      
\usepackage{amsfonts}           
\usepackage{microtype}     	

\usepackage{draftwatermark}
\SetWatermarkText{PREPRINT}
\SetWatermarkScale{1}
\definecolor{wm}{gray}{0.95}
\SetWatermarkColor{wm}

\newcommand{\papertitle}{NeuralFMU: Towards Structural Integration of FMUs into Neural Networks}

\newcommand{\mycode}[1]{{\small\texttt{#1}}}
\newcommand{\myurl}[1]{{\small\url{#1}}}

\newcommand{\libfmi}{{\emph{FMI.jl}}}
\newcommand{\libfmiflux}{{\emph{FMIFlux.jl}}}
\newcommand{\libflux}{{\emph{Flux.jl}}}
\newcommand{\libzygote}{{\emph{Zygote.jl}}}
\newcommand{\libdiffeq}{{\emph{DifferentialEquations.jl}}}
\newcommand{\libdiffeqflux}{{\emph{DiffEqFlux.jl}}}
\newcommand{\libmodia}{{\emph{Modia.jl}}}

\newcommand{\urlfmi}{{\myurl{https://github.com/ThummeTo/FMI.jl}}}
\newcommand{\urlfmiflux}{{\myurl{https://github.com/ThummeTo/FMIFlux.jl}}}
\newcommand{\urlzygote}{\myurl{https://fluxml.ai/Zygote.jl/latest/}}
\newcommand{\urlflux}{\myurl{https://fluxml.ai/Flux.jl/stable/}}
\newcommand{\urldiffeq}{\myurl{https://diffeq.sciml.ai/stable/}}
\newcommand{\urldiffeqflux}{\myurl{https://github.com/SciML/DiffEqFlux.jl}}
\newcommand{\urlmodia}{\myurl{https://github.com/ModiaSim/Modia.jl}}

\usepackage{acronym}
\newacro{FMU}[FMU]{Functional Mock-up Unit}
\newacroplural{FMU}[FMUs]{Functional Mock-up Units}
\newacro{FMI}[FMI]{Functional Mock-up Interface}
\newacro{SSP}[SSP]{System Structure and Parameterization}
\newacroplural{SSP}[SSPs]{System Structure and Parameterization}
\newacro{NN}[NN]{Neural Network}
\newacroplural{NN}[NNs]{Neural Networks}
\newacro{AD}[AD]{Automatic Differentiation}
\newacro{PINN}[PINN]{Physics-informed Neural Network}
\newacroplural{PINNs}[PINNs]{Physics-informed Neural Networks}
\newacro{CS}[CS]{co-simulation}
\newacro{ME}[ME]{model exchange}
\newacro{ODE}[ODE]{ordinary differential equation}
\newacroplural{ODE}[ODEs]{ordinary differential equations}
\newacro{BNSDE}[BNSDE]{Bayesian Neural Stochastic Differential Equation}
\newacroplural{BNSDE}[BNSDE]{Bayesian Neural Stochastic Differential Equations}
\newacro{PAC}[PAC]{Probably Approximately Correct}
\newacro{MSL}[MSL]{Modelica Standard Library}

\renewcommand{\headeright}{Preprint}
\renewcommand{\undertitle}{Preprint}
\renewcommand{\shorttitle}{~}

\hypersetup{%
  pdftitle  = {\papertitle},
  pdfauthor = {Tobias Thummerer, Josef Kircher, Lars Mikelsons},
  pdfkeywords = {NeuralFMU, FMI, FMU, Julia, NeuralODE}
}

\begin{document}

\title{\papertitle}
\author{%
	Tobias Thummerer \\
	Chair of Mechatronics\\
	Augsburg University \\
	\texttt{tobias.thummerer@informatik}\\
	\texttt{.uni-augsburg.de}  
	\And 
	Josef Kircher \\
	Chair of Mechatronics\\
	Augsburg University \\
	\texttt{josef.kircher@student}\\
	\texttt{.uni-augsburg.de}  
	\And 
	Lars Mikelsons \\
	Chair of Mechatronics\\
	Augsburg University \\
	\texttt{lars.mikelsons@informatik}\\
	\texttt{.uni-augsburg.de}  
}
\date{}
\maketitle

\begin{abstract}
This paper covers two major subjects: First, the presentation of a new open-source library called \libfmi{} for integrating \acs{FMI} into the Julia programming environment by providing the possibility to load, parameterize and simulate \acsp{FMU}. Further, an extension to this library called \libfmiflux{} is introduced, that allows the integration of \acsp{FMU} into a neural network topology to obtain a \emph{NeuralFMU}. This structural combination of an industry typical black-box model and a data-driven machine learning model combines the different advantages of both modeling approaches in one single development environment. This allows for the usage of advanced data driven modeling techniques for physical effects that are difficult to model based on first principles.
\end{abstract}


\begin{multicols}{2}

\section{Introduction}
Models inside closed simulation tools make hybrid modeling difficult, because for training data-driven model parts, determination of the loss gradient through the \ac{NN} and the model itself is needed. Nevertheless, the structural integration of models inside machine learning topologies like \acp{NN} is a research topic that gathered more and more attention. When it comes to learning system dynamics, the structural injection of algorithmic numerical solvers into \acp{NN} lead to large improvements in performance, memory cost and numerical precision over the use of residual neural networks \cite{Chen:2018}, while offering a new range of possibilities, e.g. fitting data that was observed at irregular time steps \cite{Innes:2019}. The result of integrating a numerical solver for \acp{ODE} into a \ac{NN} is known as \emph{NeuralODE}. For the Julia programming language (from here on simply referred to as \emph{Julia}), a ready-to-use library for building and training NeuralODEs named \libdiffeqflux{}\footnote{\urldiffeqflux{}} is already available \cite{Rackauckas:2019}. Probably the most mentioned point of criticism regarding NeuralODEs is the difficult porting to real world applications (s. section \ref{sec:method_fw} and \ref{sec:method_adjoint}).

A different approach for hybrid modeling, as in \textcite{Raissi:2019}, is the integration of the physical model into the machine learning process by evaluating (forward propagating) the physical model as part of the loss function during training in so called \acp{PINN}. In contrast, this contribution focuses on the structural integration of \acp{FMU} into the \ac{NN} itself and \emph{not} only the cost function, allowing much more flexibility with respect to what can be learned and influenced. However, it is also possible to build and train \acp{PINN} with the presented library. 

Finally, another approach are \acp{BNSDE} as is \textcite{Haussmann:2021}, which use bayesian model selection together with \ac{PAC} bayesian bounds during the \ac{NN} training to improve hybrid model accuracy on basis of noisy prior knowledge. For an overview on the growing field of hybrid modeling see e.g. \cite{willard:2020} or \cite{rai:2020}. 

To conclude, hybrid modeling with its different facets is an emerging research field, but still chained to academic use-cases. It seems a logical next step to open up these auspicious ML-technologies, besides many more not mentioned, to industrial applications. 

\end{multicols}
\twocolumn{}
Combining physical and data-driven models inside a single industry tool is currently not possible, therefore it is necessary to port models to a more suitable environment. An industry typical model exchange format is needed. Because the \ac{FMI} is an open standard and widely used in industry as well as in research applications, it is a suitable candidate for this aim. Finally, a software interface that integrates \ac{FMI} into the ML-environment is necessary. Therefore, we present two open-source software libraries, which offer all required features:
\begin{itemize}
	\item \libfmi{}: load, instantiate, parameterize and simulate \acp{FMU} seamlessly inside the Julia programming language
	\item \libfmiflux{}: place \acp{FMU} simply inside any feed-forward \ac{NN} topology and still keep the resulting hybrid model trainable with a standard \ac{AD} training process 
\end{itemize}
Because the result of integrating a numerical solver into a \ac{NN} is known as \emph{NeuralODE}, we suggest to pursue this naming convention by presenting the integration of a \ac{FMU}, \ac{NN} and a numerical solver as \emph{NeuralFMU}. 

By providing the libraries \libfmi{} (\urlfmi{}) and \libfmiflux{} (\urlfmiflux{}), we want to open the topic \emph{NeuralODEs} for industrial applications, but also lay a foundation to bring other state-of-the-art ML-technologies closer to production.
In the following two subsections, short style explanations of the involved tools and techniques are given.

\subsection{Julia Programming Language}
In this section, it is shortly explained and motivated why the authors picked the Julia programming language for the presented task. Julia is a dynamic typing language developing since 2009 and first published in 2012 \cite{Bezanson:2012}, with the aim to provide fast numerical computations in a platform-independent, high-level programming language \cite{Bezanson:2015}. The language and interpreter was originally invented at the \emph{Massachusetts Institute of Technology}, but since today many other universities and research facilities joined the development of language expansions, which mirrors in many contributions from different countries and even in its own conference, the \emph{JuliaCon}\footnote{\myurl{http://www.juliacon.org}}.
In \textcite{Elmqvist:2018}, the library expansion \libmodia{}\footnote{\urlmodia{}} was introduced. \libmodia{} allows object-orientated white-box modeling of mechanical and electrical systems, syntactically similar to \emph{Modelica}, in Julia. 

\subsection{\acl{FMI} (\acs{FMI})}
The \ac{FMI}-standard \cite{FMI:2020} allows the distribution and exchange of models in a standardized format and independent of the modeling tool. An exported model container, that fulfills the \ac{FMI}-requirements is called \ac{FMU}. \acp{FMU} can be used in other simulation environments or even inside of entire \aclp{CS} like \ac{SSP} \cite{SSP:2019}. \acp{FMU} are subdivided into two major classes: \ac{ME} and \ac{CS}. The different use-cases depend on the \ac{FMU}-type and the availability of standardized, but optional implemented \ac{FMI}-functions.

This paper is further structured into four topics: The presentation of our libraries \libfmi{} and \libfmiflux{}, an example handling a NeuralFMU setup and training, the explanation of the methodical background and finally a short conclusion with future outlook.

\section{Presenting the Libraries}
Our Julia-library \libfmi{} provides high-level commands to unzip, allocate, parameterize and simulate entire \acp{FMU}, as well as plotting the solution and parsing model meta data from the model description. Because \ac{FMI} has already two released specification versions and is under ongoing development\footnote{The current version is 2.0.2, but an alpha version 3.0 is already available.}, one major goal was to provide the ability to simulate different version \acp{FMU} with the same user front-end. To satisfy users who prefer close-to-specification programming, as well as users that are new to the topic and favor a smaller but more high-level command set, we provide high-level Julia commands, but also the possibility to use the more low-level commands specified in the \ac{FMI}-standard \cite{FMI:2020}.

\subsection{Simulating FMUs}
\begin{figure}[h!]
	\centering
	\includegraphics[width=0.4 \textwidth,trim={0 1.5cm 0 1.5cm},clip]{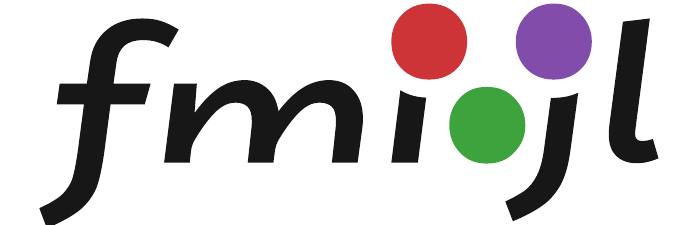}
	\caption{Logo of the library \libfmi{}.}
	\label{fig:logo_fmijl}
\end{figure}
The shortest way to load a \ac{FMU} with \libfmi{}, simulate it for $t \in \{0,10\}$, gather and plot simulation data for the example variable \emph{mass.s} and free the allocated memory is implemented only by a few lines of code as follows:
\begin{lstlisting}[caption={Simulating \ac{FMU}s with \libfmi{} (high-level).}, label={lst:simfmu}]
using FMI
myFMU = fmiLoad("path/to/myFMU.fmu")
fmiInstantiate!(myFMU)
simData = fmiSimulate(myFMU, 0.0, 10.0; recordValues=["mass.s"])
fmiPlot(simData)
fmiUnload(myFMU)		
\end{lstlisting}
Please note, that these six lines are not only a code snippet, but a fully runnable Julia program.

For users, that prefer more control over memory and performance, the original C-language command set from the \ac{FMI}-specification is wrapped into low-level commands and is available, too. A code snippet, that simulates a \ac{CS}-\ac{FMU}, looks like this:
\begin{lstlisting}[caption=Simulating \acs{CS}-\acp{FMU} with \libfmi{} (low-level).,label=lst:simfmu_low]
using FMI
myFMU = fmiLoad("path/to/myFMU.fmu")
fmuComp = fmiInstantiate!(myFMU)
fmiSetupExperiment(fmuComp, 0.0, 10.0)
fmiEnterInitializationMode(fmuComp)
fmiExitInitializationMode(fmuComp)
dt = 0.1
ts = 0.0:dt:10.0
for t in ts
  fmiDoStep(fmuComp, t, dt)
end
fmiTerminate(fmuComp)
fmiFreeInstance!(fmuComp)
fmiUnload(myFMU)			
\end{lstlisting}
Note, that these function calls are not dependent on the \ac{FMU}-Version, but are inspired by the command set of \ac{FMI} 2.0.2. The underlying \ac{FMI}-Version is determined in the call \mycode{fmiLoad}. Because the naming convention could change in future versions of the standard, version-specific function calls like \mycode{fmi2DoStep} (the "2" stands for the \ac{FMI}-versions 2.x) are available, too. Readers that are familiar with \ac{FMI} will notice, that the functions \mycode{fmiLoad} and \mycode{fmiUnload} are not mentioned in the standard definition. The function \mycode{fmiLoad} handles the creation of a temporary directory for the unpacked data, unpacking of the \ac{FMU}-archive, as well as the loading of the \ac{FMU}-binary and its model description. In \mycode{fmiUnload}, all \ac{FMU}-related memory is freed and the temporary directory is deleted. Beside \ac{CS}-\acp{FMU}, \ac{ME}-\acp{FMU} are supported, too. The numerical solving and event handling for \ac{ME}-\acp{FMU} is performed via the library \libdiffeq{}\footnote{\urldiffeq{}}, the standard library for solving different types of differential equations in Julia \cite{Rackauckas:2021}.

\subsection{Integrating \acp{FMU} into \acp{NN}}
\begin{figure}[h!]
	\centering
	\includegraphics[width=0.4 \textwidth,trim={0 1.0cm 0 1.0cm},clip]{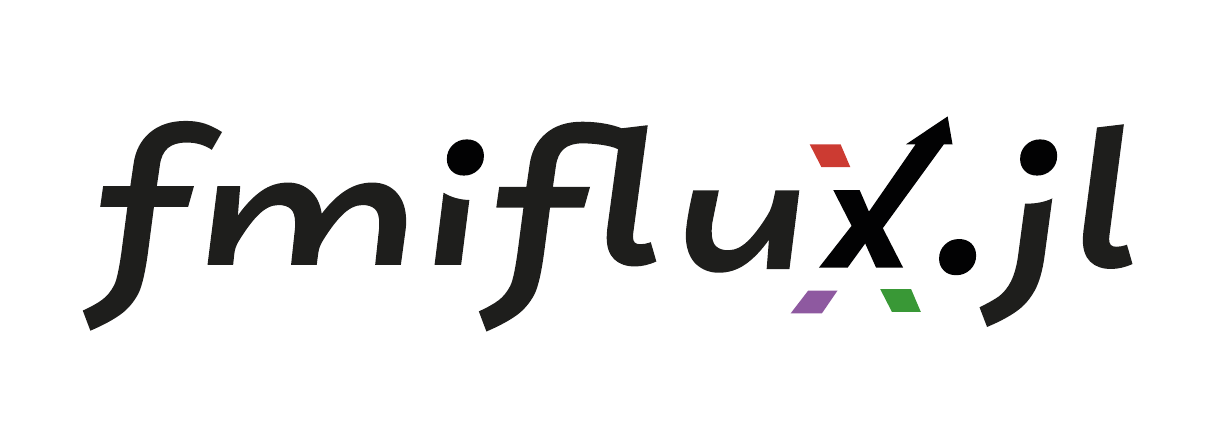}
	\caption{Logo of the library extension \libfmiflux{}.}
	\label{fig:logo_fmifluxjl}
\end{figure}
The open-source library extension \libfmiflux{} allows for the fusion of a \ac{FMU} and a \ac{NN}. As in many other machine learning frameworks, a deep \ac{NN} in Julia using \libflux{}\footnote{\urlflux{}} is configured by chaining multiple neural layers together. Probably the most intuitive way of integrating a \ac{FMU} into this topology, is to simply handle the \ac{FMU} as a network layer. In general, \libfmiflux{} does not make restrictions to ...
\begin{itemize}
	\item ... which \ac{FMU}-signals can be used as layer inputs and outputs. It is possible to use any variable that can be set via \mycode{fmiSetReal} or \mycode{fmiSetContinuousStates} as input and any variable that can be retrieved by \mycode{fmiGetReal} or \mycode{fmiGetDerivatives} as output.
	\item ... where to place \acp{FMU} inside the \ac{NN} topology, as long as all signals are traceable via \ac{AD} (no signal cuts).
\end{itemize}
Dependent on the \ac{FMU}-type, \ac{ME} or \ac{CS}, different setups for NeuralFMUs should be considered. In the following, two possibilities are presented.

\subsubsection{\ac{ME}-\acp{FMU}}\label{sec:me_fmus}

For most common applications, the use of \ac{ME}-\acp{FMU} will be the first choice. Because of the absence of an integrated numerical solver inside the \ac{FMU}, there are much more possibilities when it comes to learning dynamic processes. A mathematical view on a \ac{ME}-\ac{FMU} leads to the state space equation (\autoref{eq:me_ss}) and output equation (\autoref{eq:me_out}), meaning a \ac{ME}-\ac{FMU} computes the state derivative $\dot{\vec{x}}_{me}$ and output values $\vec{y}_{me}$ for the current time step $t$ from a given state $\vec{x}_{me}$ and optional input $\vec{u}_{me}$:
\begin{align}
\dot{\vec{x}}_{me} = \vec{f}_{me}(\vec{x}_{me}, \vec{u}_{me}, t) \label{eq:me_ss}\\
\vec{y}_{me} = \vec{g}_{me}(\vec{x}_{me}, \vec{u}_{me}, t) \label{eq:me_out}
\end{align}

\begin{figure}[h!]
	\centering
	\includegraphics[width=0.34 \textwidth]{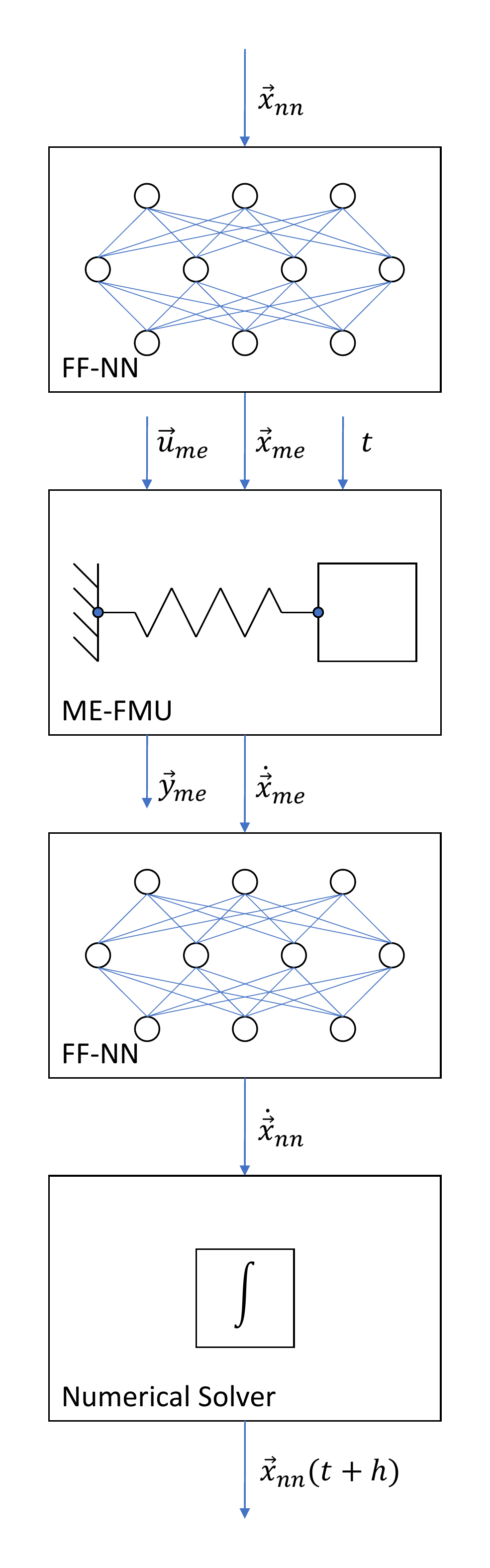}
	\caption{Example for a Neural\ac{FMU} (\ac{ME}).}
	\label{fig:example_me}
	\vspace{0.5cm}
\end{figure} 

Different interfaces between the \ac{FMU} layer and \ac{NN} are possible. For example, the number of \ac{FMU} layer inputs could equal the number of \ac{FMU} layer outputs and be simply the number of model states. In \autoref{fig:example_me}, the visualization of the suggested structure is given. The top \ac{NN} is fed by the current system state $\vec{x}_{nn}$. The \ac{NN} is able to learn and compensate state-dependent modeling failures like measurement offsets or physical displacements and thresholds. After that, the corrected state vector $\vec{x}_{me}$ is passed to the \ac{ME}-\ac{FMU}, and the current state derivative $\dot{\vec{x}}_{me}$ is retrieved. The bottom \ac{NN} is able to learn additional physical effects, like friction or other forces, from the state derivative vector. Finally the corrected state derivatives $\dot{\vec{x}}_{nn}$ are integrated by the numerical solver, to retrieve the next system state $\vec{x}_{nn}(t+h)$. Note, that the time step size $h$ can be determined by a modern numerical solver like \emph{Tsit5} \cite{Tsitouras:2011}, with different advantages like dynamic step size and order adaption. This is a significant advantage in performance and memory cost over the use of recurrent \acp{NN} for numerical integration.

Note, that many other configurations for setting up the NeuralFMU are thinkable, e.g.:
\begin{itemize}
	\item the top \ac{NN} could additionally generate a \ac{FMU} input $\vec{u}_{me}$
	\item the bottom \ac{NN} could learn from states $\vec{x}_{me}$ or derivatives $\dot{\vec{x}}_{me}$, for a targeted expansion of the model by additional model equations
	\item the bottom \ac{NN} could learn from the \ac{FMU} output $\vec{y}_{me}$ or other model variables, that can be retrieved by \mycode{fmiGetReal}
	\item there could be a bypass around the \ac{FMU} between both \acp{NN} to forward state-dependent signals from the top \ac{NN} to the bottom \ac{NN}
	\item of course, there is no restriction to fully-connected (dense) layers, other feed-forward layers or even drop-outs are possible
\end{itemize}

To implement the presented \ac{ME}-Neural\ac{FMU} in Julia, the following code is sufficient:
\begin{lstlisting}[caption=A Neural\ac{FMU} (\ac{ME}) in Julia.,label=lst:chain_me]
net = Chain(
 Dense(length(x_nn), ...), 
 ...,
 Dense(..., length(x_me)),
 x_me -> fmiDoStepME(myFMU, x_me),
 Dense(length(dx_me), ...), 
 ...,
 Dense(..., length(dx_nn)))
nfmu = ME_NeuralFMU(net, ...)
\end{lstlisting}

\subsubsection{\ac{CS}-\acp{FMU}}\label{sec:cs_fmus}
\begin{figure}[h!]
	\centering
	\includegraphics[width=0.34 \textwidth]{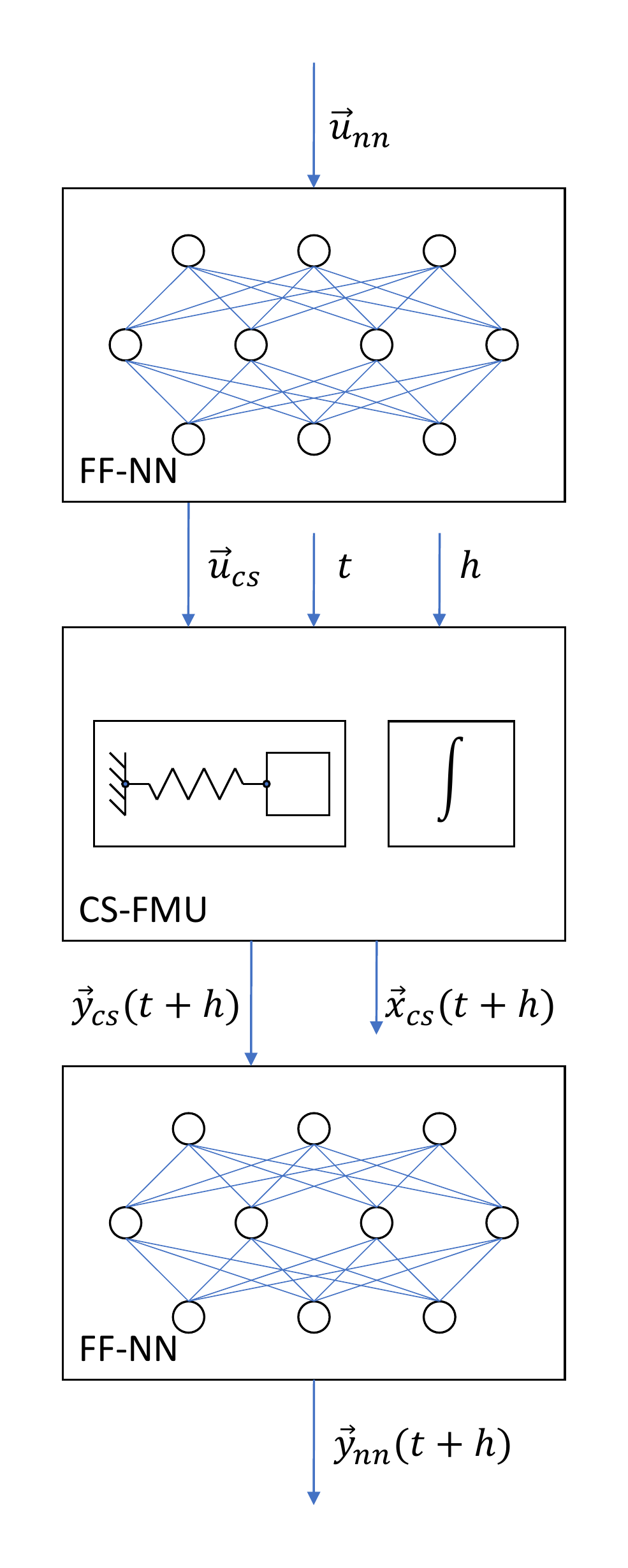}
	\caption{Example for integrating a \ac{CS}-\ac{FMU} into a \ac{NN}.}
	\label{fig:example_cs}
\end{figure}
Beside \ac{ME}-\acp{FMU}, it is also possible to use \ac{CS}-\acp{FMU} as part of NeuralFMUs. For \ac{CS}-\acp{FMU}, a numerical solver like \emph{CVODE} \cite{CVODE:2021} is already integrated and compiled as part of the \ac{FMU} itself. 

This prevents the manipulation of system dynamics at the most effective point: Between the \ac{FMU} state derivative output and the numerical integration. However, other tasks like learning an error correction term, are still possible to implement. The presence of a numerical solver leads to a different mathematical description compared to \ac{ME}-\acp{FMU}: The \ac{CS}-\ac{FMU} computes the next state $\vec{x}_{cs}(t+h)$ and output $\vec{y}_{cs}(t+h)$ dependent on its internal current state $\vec{x}_{cs}$ (Eq. \ref{eq:cs_ss} and \ref{eq:cs_out}). Unlike for \ac{ME}, the state and derivative values of a \ac{CS}-\ac{FMU} are not necessarily known (disclosed via \ac{FMI}). 
\begin{align}
\vec{x}_{cs}(t+h) = \vec{f}_{cs}(\vec{x}_{cs}, \vec{u}_{cs}, t, h) \label{eq:cs_ss}\\
\vec{y}_{cs}(t+h) = \vec{g}_{cs}(\vec{x}_{cs}, \vec{u}_{cs}, t, h)  \label{eq:cs_out}
\end{align}
In case of \ac{CS}-\acp{FMU}, the number of layer inputs could be based on the number of \ac{FMU}-inputs and the number of outputs in analogy. As for \ac{ME}-NeuralFMUs, this is just one possible setup for a \ac{CS}-NeuralFMU. \autoref{fig:example_cs} shows the topology of the considered Neural\ac{FMU}. The top \ac{NN} retrieves an input signal $\vec{u}_{nn}$, which is corrected into $\vec{u}_{cs}$. Note, that here training of the top \ac{NN} is only possible if the \ac{FMU} output is sensitive to the \ac{FMU} input. This is often not the case for physical simulations. Inside the \ac{CS}-\ac{FMU}, the input $\vec{u}_{cs}$ is set, an integrator step with step size $h$ is performed and the \ac{FMU} output $\vec{y}_{cs}(t+h)$ is forwarded to the bottom \ac{NN}. Here, a simple error correction into $\vec{y}_{nn}(t+h)$ is performed, meaning the error is determined and compensated without necessarily learning the mathematical representation of the underlying physical effect.

Note that for \ac{CS}, even if the macro step size $h$ must be determined by the user, it does not need to be constant if the numerical solver inside the \ac{FMU} supports varying step sizes. If so, the internal solver step size may vary from $h$, in fact $h$ acts as a upper boundary for the internal micro step size. As a result, if the \ac{FMU} is compiled with a variable-step solver, unnecessarily small values for $h$ will have negative influence on the internal solver performance, but large values will not destabilize the numerical integration.

Other configurations for setting up the hybrid structure are interesting, e.g.:
\begin{itemize}
	\item the bottom \ac{NN} could learn from the system state $\vec{x}_{cs}(t+h)$ or state derivative $\dot{\vec{x}}_{cs}(t+h)$
	\item the step size $h$ could be learned by an additional \ac{NN} to optimize simulation performance
	\item there could be a bypass around the \ac{FMU} between both \acp{NN} to forward input-dependent signals from the top \ac{NN} to the bottom \ac{NN}
\end{itemize}

The software implementation of the considered \ac{CS}-Neural\ac{FMU} looks as follows:
\begin{lstlisting}[caption=A Neural\ac{FMU} (\ac{CS}) in Julia.,label=lst:chain_cs]
net = Chain(
 Dense(length(u_nn), ...), 
 ...,
 Dense(..., length(u_cs)),
 u_cs -> fmiInputDoStepCSOutput(myFMU, h, u_cs),
 Dense(length(y_cs), ...), 
 ...,
 Dense(..., length(y_nn)))
nfmu = CS_NeuralFMU(net, ...)
\end{lstlisting}
First, the function \mycode{fmiInputDoStepCSOutput} sets all \ac{FMU}-inputs to the values \mycode{u}, respectively the output of the previous net layer. After that, a \mycode{fmiDoStep} with step size \mycode{h} is performed and finally the \ac{FMU} output is retrieved and passed to the next layer. Because of the integration over time inside the \ac{CS}-\ac{FMU} to retrieve new system states, it is necessary to reset the \ac{FMU} for every training run, similar to the training of recurrent \acp{NN}.

\section{Methodical Background}\label{sec:method}
High-performance machine learning frameworks like \libflux{} are using \ac{AD} for reverse-mode differentiation through the \ac{NN} topology. Because all mathematical operations inside the \ac{NN} are known (white-box), this is a very efficient way to derive the gradient and train \acp{NN}. On the other hand, the jacobian over a black-box \ac{FMU} is unknown from the view of an \ac{AD} framework, because the model structure is (in general) hidden as compiled machine code. This jacobian is part of the loss gradient \ac{AD}-chain and needs to be determined.

Beside others, the most common and default \ac{AD} framework in Julia is \libzygote{}\footnote{\urlzygote{}}. A remarkable feature of \libzygote{} is, that it provides the ability to define custom adjoints, meaning the gradient of a custom function can be defined to be an arbitrary function. This renders the possibility to pass a custom gradient for a \ac{FMU}-representation to the \ac{AD}-framework, which will be later used to derive the total loss function gradient during the training process.

\subsection{Gradient of the Loss Function}
For the efficient training of \acp{NN}, the gradient of the loss function according to the net parameters (weights and biases) is needed. In the following, three methods to derive the loss gradient will be discussed: \ac{AD}, forward sensitivity analysis and backward adjoints. In the following, only \ac{ME}-Neural\acp{FMU} are considered and the loss function $l(\vec{x}_{nn}(\vec{p}))$ is expected to depend explicitly on the system state $\vec{x}_{nn}$ (the NeuralFMU output) and only implicitly on the net parameters $\vec{p}$.

\subsubsection{\acl{AD} (\acs{AD})}
For white-box systems, like native implemented numerical solvers, one possible approach to provide the gradient is \ac{AD} \cite{Rackauckas:2019}. In general, the mathematical operations inside a \ac{FMU} are not known (compiled binary), meaning despite \ac{AD} being a very common technique, it is only suitable for determining the gradient of the \ac{NN}, but not the jacobian over a \ac{FMU}. The unknown jacobian $\vec{J}_{fmu}$ over the \ac{FMU} layer with layer inputs $\vec{u}$ and outputs $\vec{y}$ is noted as follows:
\begin{align}
\vec{J}_{fmu} = \frac{\partial \vec{y}}{\partial \vec{u} }
\end{align}

Inserting $\vec{u} = \vec{x}_{me}$ and $\vec{y} = \dot{\vec{x}}_{me}$ results in the jacobian $\vec{J}_{me}$ for the \ac{FMU} from the example in subsection \ref{sec:me_fmus} (\ac{ME}), inserting $\vec{u} = \vec{u}_{cs}$ and $\vec{y} = \vec{y}_{cs}(t+h)$ results in the jacobian matrix $\vec{J}_{cs}$ for subsection \ref{sec:cs_fmus} (\ac{CS}):
\begin{align}
\vec{J}_{me} &= \frac{\partial \dot{\vec{x}} }{\partial \vec{x} } \label{eq:j_me} \\
\vec{J}_{cs} &= \frac{\partial \vec{y}(t+h)}{\partial \vec{u}(t)} \approx \frac{\partial \vec{y}(t)}{\partial \vec{u}(t)} \label{eq:j_cs}
\end{align}

The simplification in \autoref{eq:j_cs} does not lead to problems for small step sizes $h$, because in practice, a small error in the jacobian only negatively affects the optimization performance (convergence speed) and not the convergence itself. However, the quantity of the mentioned error is dependent on the the optimization algorithm and parameterization and $h$ should be selected on the basis of expert knowledge about the model and optimizer or - if not available - as part of hyper parameter tuning.

\subsubsection{Forward Sensitivities}\label{sec:method_fw}
To retrieve the partial derivative (sensitivity) of the system state according to a net parameter $p_i \in \vec{p}$ and thus in straight forward manner also the gradient of the loss function, another common approach is Forward Sensitivity Analysis. Sensitivities can be estimated by extending the system state by additional sensitivity equations in form of \acp{ODE}. Dependent on the number of parameters $|\vec{p}|$, this leads to large \ac{ODE} systems of size $(1+ |\vec{p}|) \cdot |\vec{x}|$ \cite[21]{CVODES:2006} and therefore worsens the overall computation and memory performance. Computations can be reduced, but at a higher memory cost \cite[15]{Rackauckas:2019}. For a \ac{ME}-Neural\ac{FMU}, the sensitivity equation for a parameter $p_i$ can be formulated as in \textcite[19]{CVODES:2006}:
\begin{align}
\frac{d}{dt} \frac{\partial \vec{x}_{nn}}{\partial p_i} = \underbrace{\frac{\partial \dot{\vec{x}}_{nn}}{\partial \vec{x}_{nn}}}_{\vec{J}_{nn}} \cdot \frac{\partial \vec{x}_{nn}}{\partial p_i} + \frac{\partial \dot{\vec{x}}_{nn}}{\partial p_i} \label{eq:fs}
\end{align}

The jacobian of the entire Neural\ac{FMU} $\vec{J}_{nn}$ can be described via chain-rule as a product of the three jacobians $\vec{J}_{bottom}$ (over the bottom  part of the \ac{NN}), $\vec{J}_{me}$ (over the \ac{ME}-\ac{FMU}) and $\vec{J}_{top}$ (over the top part of the \ac{NN}):
\begin{align}
\underbrace{\frac{\partial \dot{\vec{x}}_{nn}}{\partial \vec{x}_{nn}}}_{\vec{J}_{nn}} = \underbrace{\frac{\partial \dot{\vec{x}}_{nn}}{\partial \dot{\vec{x}}_{me}}}_{\vec{J}_{bottom}} \cdot \underbrace{\frac{\partial \dot{\vec{x}}_{me}}{\partial \vec{x}_{me}}}_{\vec{J}_{me}} \cdot \underbrace{\frac{\partial \vec{x}_{me}}{\partial \vec{x}_{nn}}}_{{\vec{J}_{top}}} \label{eq:j_split}
\end{align}

Inserting \autoref{eq:j_split} into \autoref{eq:fs} yields:
\begin{align}
\frac{d}{dt} \frac{\partial \vec{x}_{nn}}{\partial p_i} = \underbrace{\frac{\partial \dot{\vec{x}}_{nn}}{\partial \dot{\vec{x}}_{me}}}_{\vec{J}_{bottom}} \cdot \underbrace{\frac{\partial \dot{\vec{x}}_{me}}{\partial \vec{x}_{me}}}_{\vec{J}_{me}} \cdot \underbrace{\frac{\partial \vec{x}_{me}}{\partial \vec{x}_{nn}}}_{{\vec{J}_{top}}} \cdot \underbrace{\frac{\partial \vec{x}_{nn}}{\partial p_i}}_{\vec{g}_{top\_i}} + \underbrace{\frac{\partial \dot{\vec{x}}_{nn}}{\partial p_i}}_{\vec{g}_{bottom\_i}} 
\end{align}

Retrieving the jacobian $\vec{J}_{me}$ is handled in \autoref{sec:jac}, the jacobians $\vec{J}_{bottom}$ and $\vec{J}_{top}$ are determined by \ac{AD}, because the \ac{NN} is modeled as white-box and all mathematical operations are known. The remaining gradients, $\vec{g}_{top\_i}$ and $\vec{g}_{bottom\_i}$ can be determined building an \ac{AD}-chain, dependent on the parameter locations inside the \ac{NN} (top or bottom part), the jacobian $\vec{J}_{me}$ is needed.

As mentioned, the poor scalability with parameter count makes forward sensitivities unattractive for ML-applications with large parameter spaces, but it remains an interesting option for small \acp{NN}. To decide which sensitivity approach to pick for a specific \ac{NN} size, a useful comparison according to performance of forward and other sensitivity estimation techniques, dependent on the number of involved parameters, can be found in \cite{Rackauckas:2018}.

\subsubsection{Backward Adjoints}\label{sec:method_adjoint}
The performance disadvantage of Forward Sensitivity Analysis motivates the search for a method, that scales better with a high parameter count. Retrieving the directional derivatives over a black-box \ac{FMU} sounds similar to the reverse-mode differentiation over a black-box numerical solver as in \textcite{Chen:2018}. The name \emph{Backward Adjoints} results from solving the \ac{ODE} adjoint problem backwards in time:
\begin{align}
\vec{a} &= \frac{d l(\vec{x}_{nn})}{d \vec{x}_{nn}}\\
\frac{d\vec{a}}{dt} &= -\vec{a}^T \cdot \underbrace{\frac{\partial \dot{\vec{x}}_{nn}}{\partial \vec{x}_{nn}}}_{\vec{J}_{nn}}
\end{align}
The jacobian $\vec{J}_{nn}$ can be retrieved like in \autoref{eq:j_split}.
The searched gradient of the loss function is then given as in \textcite[22]{CVODES:2006}:
\begin{align}
\frac{d l(\vec{x}_{nn})}{d\vec{p}} = \vec{a}^T(t_0) \cdot \underbrace{\frac{\partial \vec{x}_{nn}}{\partial \vec{p}}}_{\vec{g}_{top}}(t_0) + \int_{t_0}^{t_1} \vec{a}^T(t) \underbrace{\frac{\partial \dot{\vec{x}}_{nn}}{\partial \vec{p}}(t)}_{\vec{g}_{bottom}} dt
\end{align}

Here $\vec{g}_{top}$ and $\vec{g}_{bottom}$ can be determined again using \ac{AD} and $\vec{J}_{me}$.
To conclude, the backward adjoint \ac{ODE} system with dimension $|\vec{x}|$ has to be solved only once independent of the number of parameters and therefore requires less computations for large parameter spaces compared to forward sensitivities. On the other hand, backward adjoints are only suitable, if the loss function gradient is smooth and bounded \cite[22]{CVODES:2006}, which limits the possible use for this technique to continuous systems and therefore to almost only research applications.

\subsection{Jacobian of the \ac{FMU}}\label{sec:jac}
Independent of the chosen method, the jacobian over the \ac{FMU} $\vec{J}_{me}$ is needed to keep the NeuralFMU trainable, but is unknown and must be determined. In the following, we suggest two possibilities to retrieve the gradient over a \ac{FMU}: Finite Differences and the use of the built-in function \mycode{fmi2GetDirectionalDerivative}.

\subsubsection{Finite Differences}
The jacobian can be derived by selective input modification, sampling of additional trajectory points and estimating the partial derivatives via finite differences. Note, that this approach is an option for \ac{ME}-\acp{FMU}, for \ac{CS}-\acp{FMU} only if the optional functions to store and set previous \ac{FMU} states, \mycode{fmi2GetState} and \mycode{fmi2SetState}, are available. Otherwise, sampling would require to setup a new simulation for every \ac{FMU} layer input and every considered time step, if the system state vector is unknown. This would be an unacceptable effort for most industrial applications with large models. 

\subsubsection{Built-in Directional Derivatives}\label{sec:build-in-dd}
The preferred approach in this paper is different and benefits from a major advantage of the \ac{FMI}-standard: Fully implemented \acp{FMU} provide the partial derivatives between any variable pair, thus the partial derivative between the systems states and derivatives (\ac{ME}) or the \ac{FMU} inputs and its outputs (\ac{CS}) is known at any simulation time step and does \emph{not} need to be estimated by additional methods. In \ac{FMI} 2.0.2, the partial derivatives can be retrieved by calling the optional function \mycode{fmi2GetDirectionalDerivative} \cite[26]{FMI:2020}. Depending on the underlying implementation of this function, which can vary between exporting tools, this can be a fast and reliable way to gather directional derivatives in fully implemented \acp{FMU}.

To conclude, the key step is to forward the directional derivatives over the \ac{FMU} to the \ac{AD}-framework \libzygote{}. As mentioned, \libzygote{} provides a feature to define a custom gradient over any function. In this case, the gradients for the functions \mycode{fmiDoStepME} and \mycode{fmiInputDoStepCSOutput} are wrapped to calls to \mycode{fmi2GetDirectionalDerivative}. 

Finally, we provide a seamless link to the ML-library \libflux{}, meaning Neural\acp{FMU} can be trained the same way as a convenient \ac{NN} in Julia:
\begin{lstlisting}[breaklines=true,caption=Training Neural\acp{FMU} in Julia.,label=lst:train]
nfmu = NeuralFMU(net, ...)
p_net = Flux.params(nfmu)
Flux.train!(..., p_net, ...)
\end{lstlisting}

As a final note, the presented methodical procedure, integrating \acp{FMU} into the Julia machine learning environment, can be transferred to other AD-frameworks in other programming languages like \emph{Python}.

\section{Example}
When modeling physical systems, it's often not practical to model solely based on first principle and parameterize every physical aspect. For example, when modeling mechanical, electrical or hydraulic systems, a typical modeling assumption is the negligence of friction or the use of greatly simplified friction models. Even when using friction models, the parameterization of these is a difficult and error prone task. Therefore, we decided to show the benefits of the presented hybrid modeling technique on an easy to understand example from this set of problems.

\subsection{Model}
\begin{figure}[h!]
	\centering
	\includegraphics[width=0.5 \textwidth,trim={16cm0 8.5cm 16cm 9cm},clip]{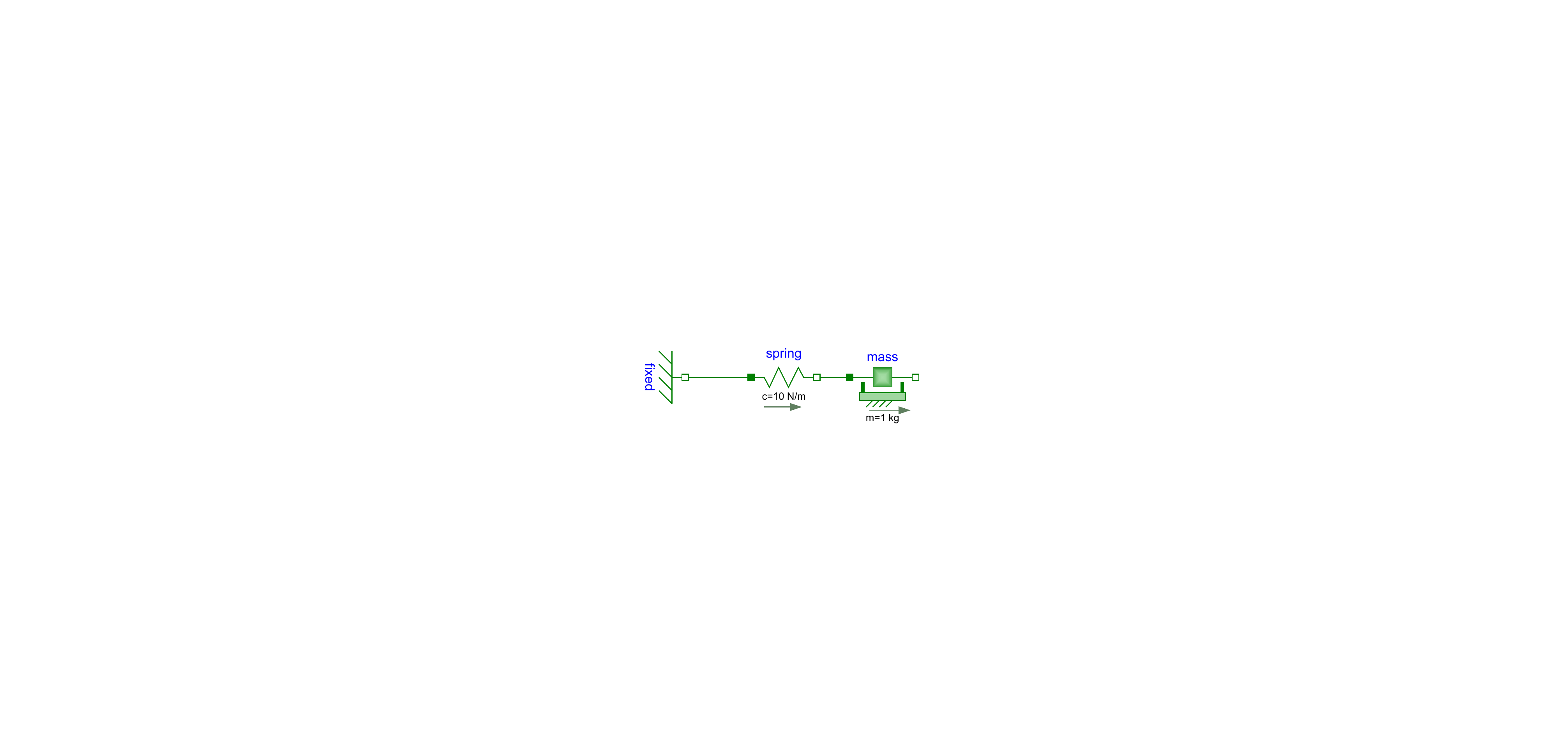}
	\caption{The reference system in \emph{Modelica}.}
	\label{fig:model}
\end{figure}
As in \autoref{fig:model}, the reference system is modeled as one mass oscillator (horizontal spring-pendulum) with mass $m$, spring constant $c$ and relaxed spring length $s_{rel}$, defined by the differential equation:
\begin{align}
\ddot{s} = \dot{v} = a = \frac{c \cdot (s_0 + s_{rel}-s) - f_{fric}(v)}{m}
\end{align}

The parameter $s_0$ describes the absolute position of the fixed anchor point, allowing to model a system displacement or a constant position measurement offset. Further, the friction force $f_{fric}$ between the pendulum body and the underlying ground is implemented with the non-linear, discontinuous friction model from \emph{MassWithStopAndFriction}\footnote{Modelica.Mechanics.Translational.Components.\allowbreak MassWithStopAndFriction (\acs{MSL} 3.2.3)} as part of the \ac{MSL}. The friction term for positive $v$ denotes:
\begin{align}
f_{fric}(v) = f_{coulomb} + f_{prop} \cdot v + f_{stribeck} \cdot e^{-f_{exp} \cdot v} \label{eq:fric}
\end{align}
This friction model consists of parameters for the constant Coulomb-friction $f_{coulomb}$, $f_{prop}$ for the velocity-proportional (viscous) friction and $f_{stribeck}$ for the exponential Stribeck-friction. The \ac{FMU} (white box) model on the other hand, only covers the modeling of a continuous, frictionless spring pendulum, therefore with $f_{fric}(v)=0$. 

The aim here is to learn a generalized representation of the parameterized friction-model in \autoref{eq:fric} from measurements of the pendulum's movement over time. Further, a displacement of $s_0=0.1\,m$ is added to the \ac{FMU} model (modeling failure), which should be detected and compensated. Both systems are parameterized as in \autoref{tab:param}.

\begin{table}[htbp]
	\caption{Parameterization of the reference and \ac{FMU} model.}\label{tab:param}
	\centering
	\begin{tabular}{p{2.3cm}rrr} \toprule
		\emph{Parameter} & \emph{Value} & \emph{Value} & \emph{Unit} \\
		~ & \emph{ref. model} & \emph{\ac{FMU} model} & ~ \\
		\midrule
		$f_{prop}$ & $0.05$ & $0.0$ & $\nicefrac{N \cdot s}{m}$ \\
		$f_{coulomb} $ & $0.25$ & $0.0$ & $N$ \\
		$f_{stribeck}$ & $0.5$ & $0.0$ & $N$ \\
		$f_{exp}$ & $2.0$ & $0.0$ & $\nicefrac{s}{m}$ \\
		$mass.m$ & $1.0$ & $1.0$ & $kg$\\
		$spring.c$ & $10.0$ & $10.0$ & $\nicefrac{N}{m}$\\
		$spring.s_{rel}$ & $1.0$ & $1.0$ & $m$\\
		$fixed.s_{0}$ & $0.0$ & $0.1$ & $m$\\
		\bottomrule
	\end{tabular}
\end{table}

\subsection{NeuralFMU Setup}
We will show, that with a \emph{NeuralFMU}-structure as in \autoref{fig:example_me}, it is possible to learn a simplified friction model as well as the constant system displacement (modeling failure) with a simple fully-connected feed-forward \ac{NN} as in \autoref{tab:nn_topo}. The network topology results from a simple random search hyper parameter optimization for a Neural\ac{FMU} model with a maximum of 150 net parameters and 8 layers. All weights are initialized with standard-normal distributed random values and all biases with zeros, except the weights of layer \#1 are initialized as identity matrix, to start training with a neutral setup and keep the system closer to the preferred intuitive solution. The loss function is defined as simple mean squared error between equidistant sample points of the Neural\ac{FMU} and the reference system.

\begin{table}[htbp]
	\caption{Topology of the example NeuralFMU.}\label{tab:nn_topo}
	\centering
	\begin{tabular}{p{3cm}rrr} \toprule
		\emph{Layer} & \emph{Inputs} & \emph{Outputs} & \emph{Activation} \\
		\midrule
		\#1 (input) & 2 & 2 & identity \\
		\#2 (FMU) & 2 & 2 & none \\
		\#3 (hidden) & 2 & 8 & identity \\
		\#4 (hidden) & 8 & 8 & tanh \\
		\#5 (output) & 8 & 2 & identity\\
		\bottomrule
	\end{tabular}
\end{table}

The corresponding code is available online as part of the library repository\footnote{\myurl{https://github.com/ThummeTo/FMIFlux.jl/blob/main/example/modelica_conference_2021.jl}}.

\subsection{Results}
\subsubsection{Training}
\begin{figure}[h!]
	\centering
	\includegraphics[width=0.5 \textwidth]{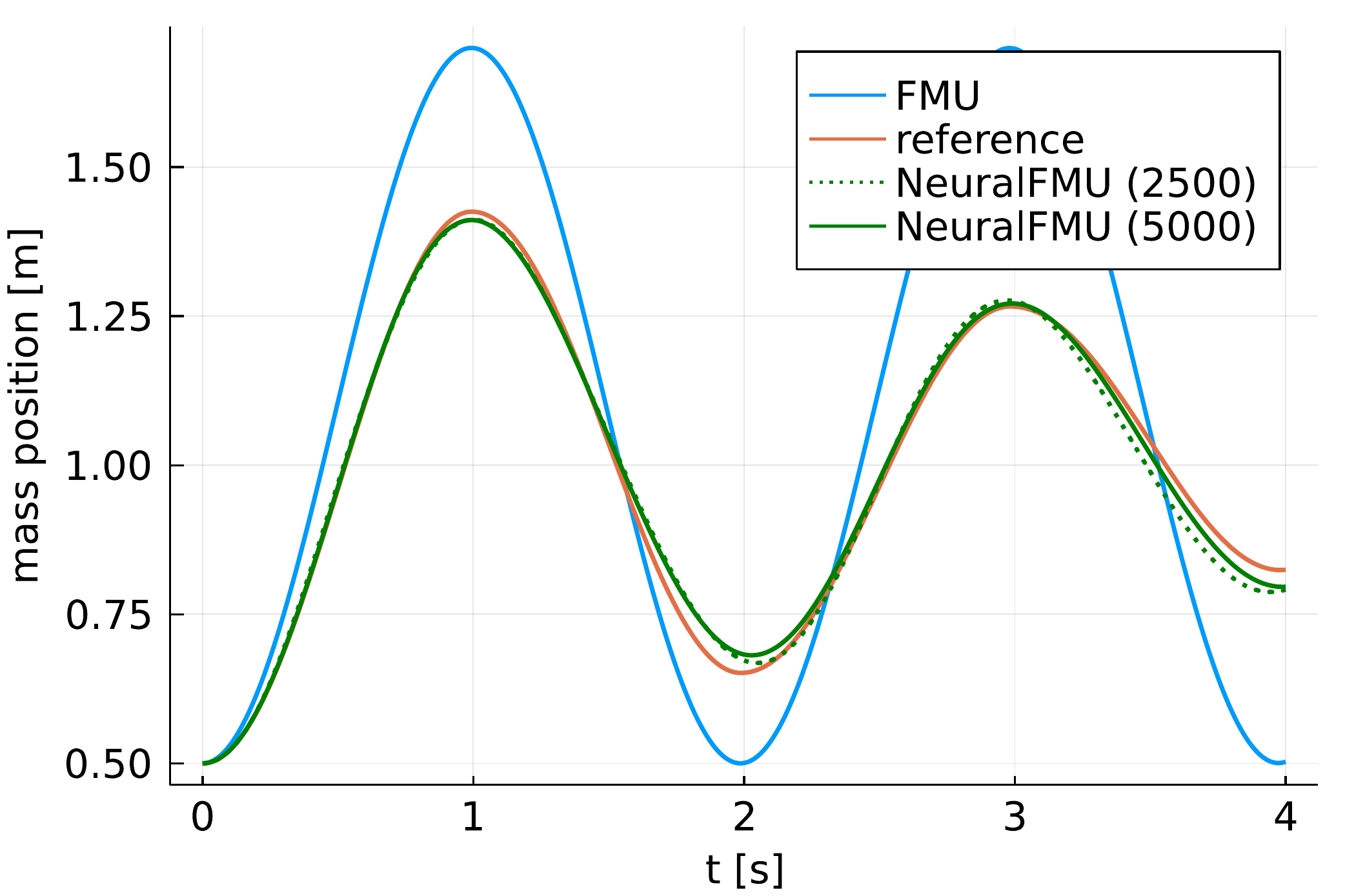}
	\caption{Comparison of the mass position of the \ac{FMU}, reference system and the NeuralFMU after 2500 and 5000 training steps on training data.}
	\label{fig:result_s_train}
\end{figure}

\begin{figure}[h!]
	\centering
	\includegraphics[width=0.5 \textwidth]{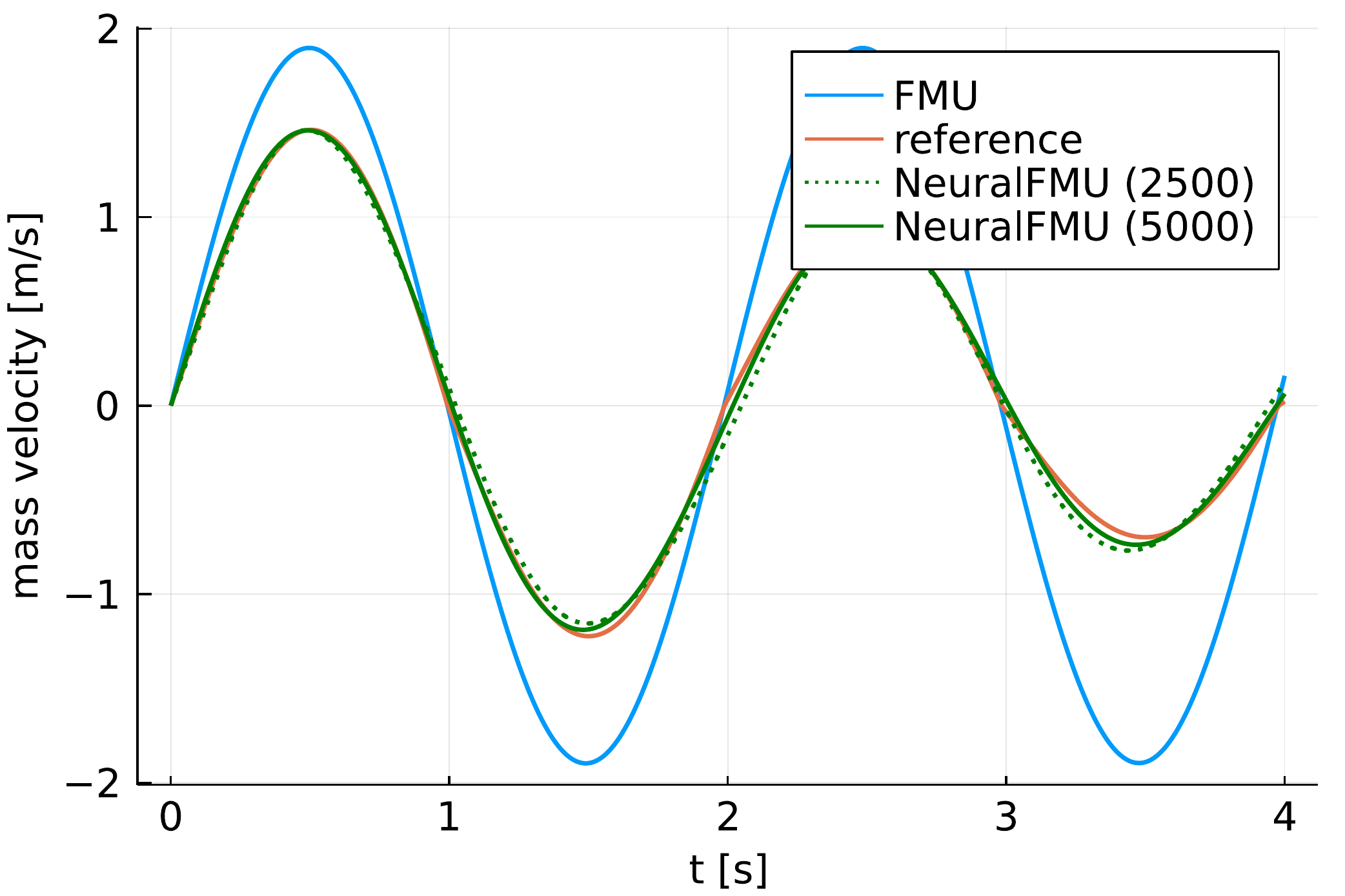}
	\caption{Comparison of the mass velocity of the \ac{FMU}, reference system and the NeuralFMU after 2500 and 5000 training steps on training data.}
	\label{fig:result_v_train}
\end{figure}

After a short training\footnote{Training was performed single-core on a desktop CPU (Intel\textregistered{} Core\textsuperscript{TM} i7-8565U) and took about 22.5 minutes. GPU training is under development.} of 2500 runs on 400 data points (each position and velocity), the hybrid model is able to imitate the reference system on training data, as can be seen in Fig. \ref{fig:result_s_train} for position and \ref{fig:result_v_train} for velocity. The training has not converged yet, further training will lead to a improved fit. For the training case, the system was initialized with $mass.s_0=0.5\,m$ (the pendulum equilibrium is at $1.0\,m$) and $mass.v_0=0\,\nicefrac{m}{s}$. Please keep in mind that the Neural\ac{FMU} was only trained by data gathered from one single simulation scenario. 


\subsubsection{Testing}
\begin{figure}[h!]
	\centering
	\includegraphics[width=0.5 \textwidth]{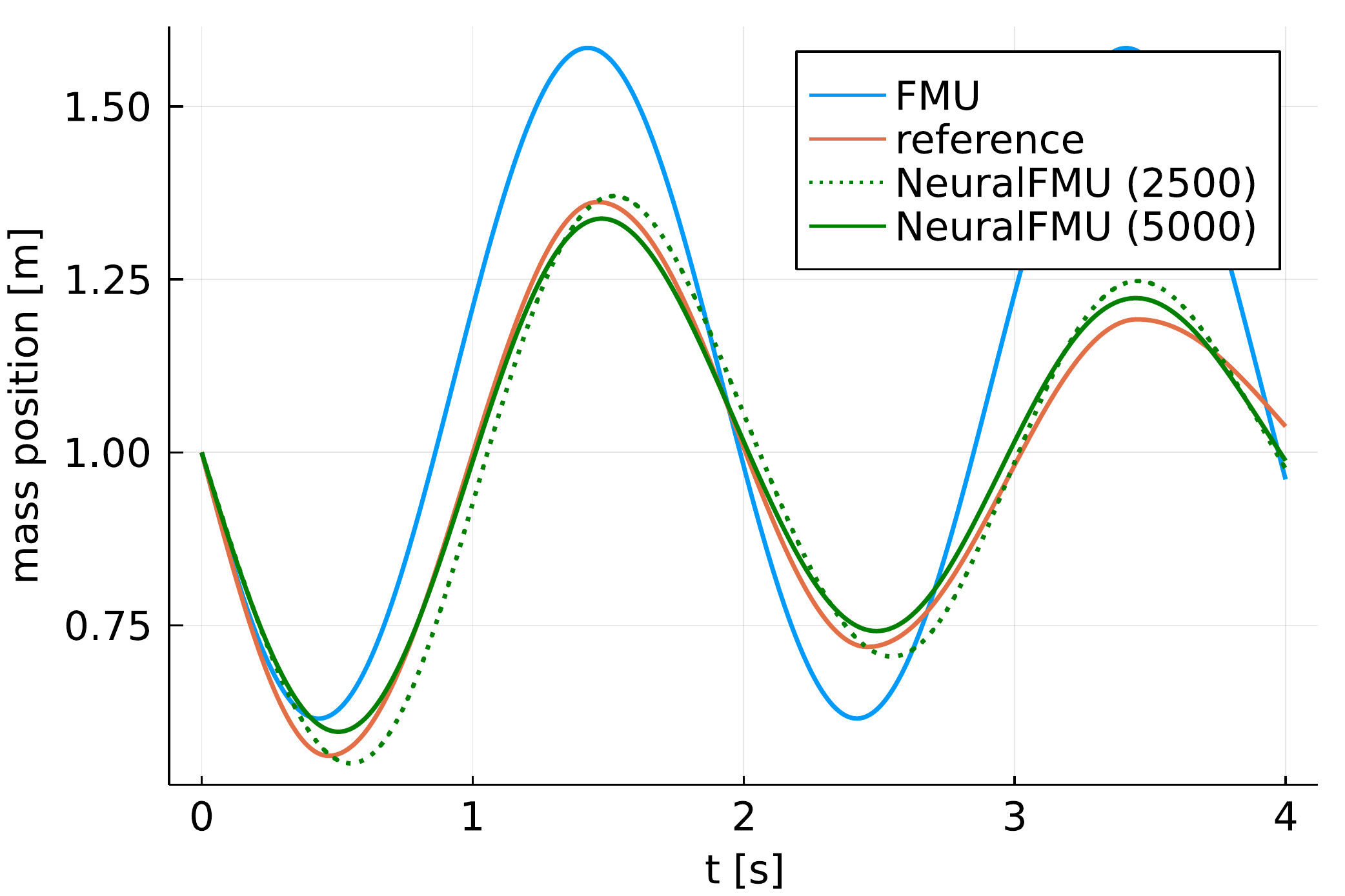}
	\caption{Comparison of the mass position of the \ac{FMU}, reference system and the NeuralFMU after 2500 and 5000 training steps on testing data.}
	\label{fig:result_s_test}
\end{figure}

\begin{figure}[h!]
	\centering
	\includegraphics[width=0.5 \textwidth]{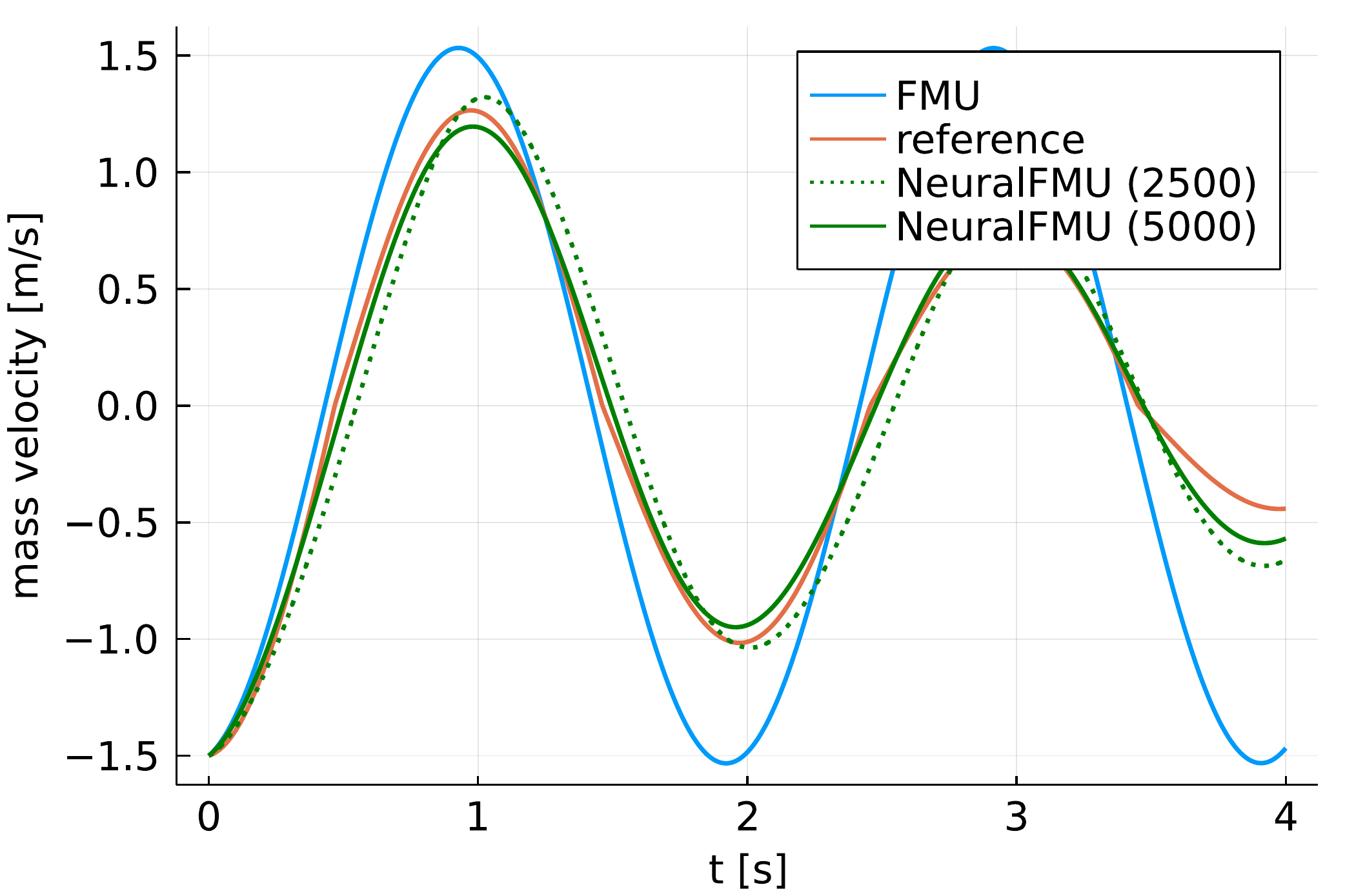}
	\caption{Comparison of the mass velocity of the \ac{FMU}, reference system and the NeuralFMU after 2500 and 5000 training steps on testing data.}
	\label{fig:result_v_test}
\end{figure}

Even if the deviation between Neural\ac{FMU} and reference system is larger for testing then for training data, the hybrid model performs well on the test case with a different (untrained) initial system state (\autoref{fig:result_s_test} and \ref{fig:result_v_test}). For testing, the system is initialized with $mass.s_0=1.0\,m$ and $mass.v_0=-1.5\,\nicefrac{m}{s}$.

\begin{figure}[h!]
	\centering
	\includegraphics[width=0.5 \textwidth]{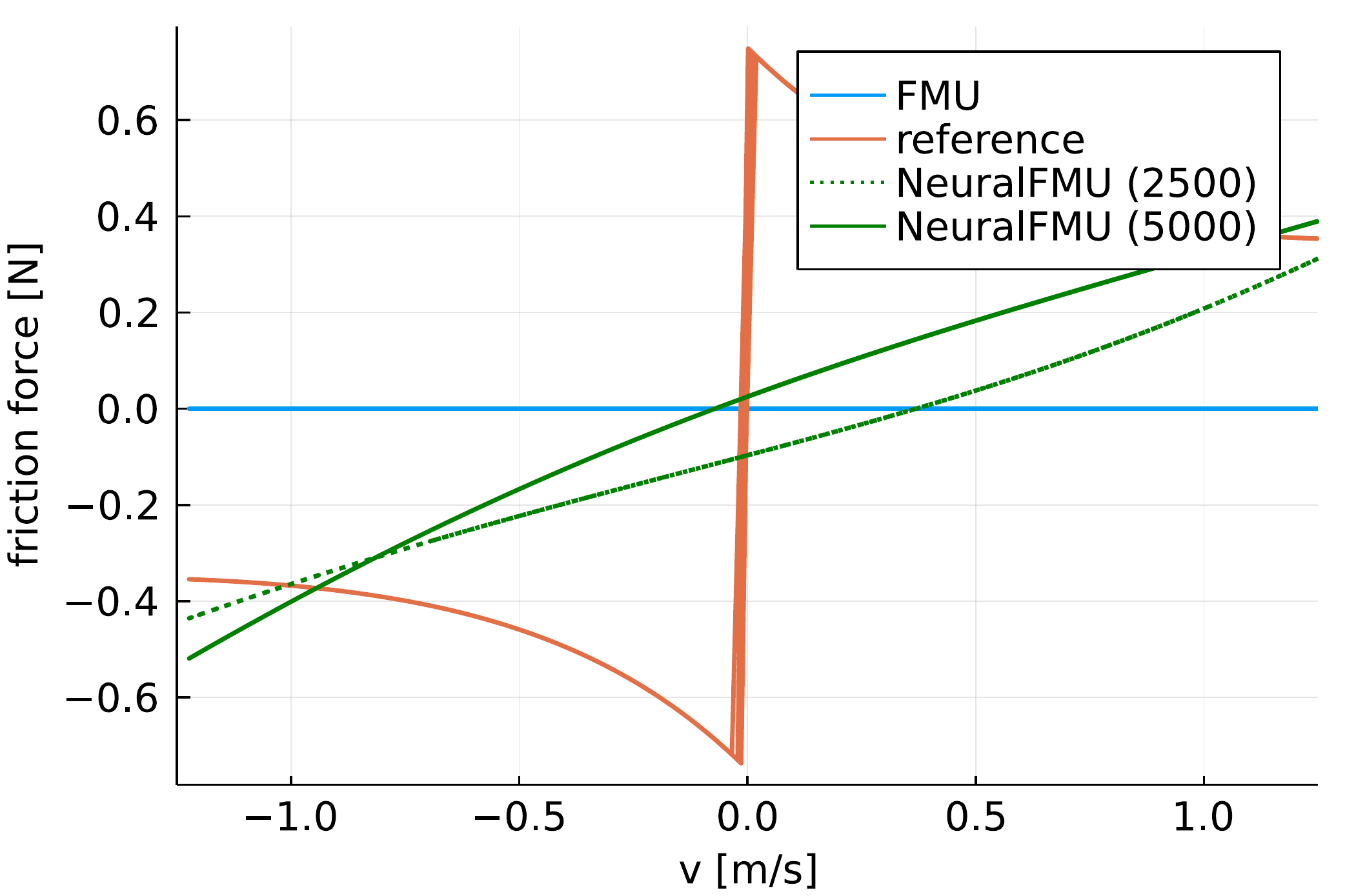}
	\caption{Comparison of the friction models of the \ac{FMU}, reference system and the NeuralFMU (bottom part \ac{NN}) after 2500 and 5000 training steps on testing data.}
	\label{fig:fric}
\end{figure}

\begin{figure}[h!]
	\centering
	\includegraphics[width=0.5 \textwidth]{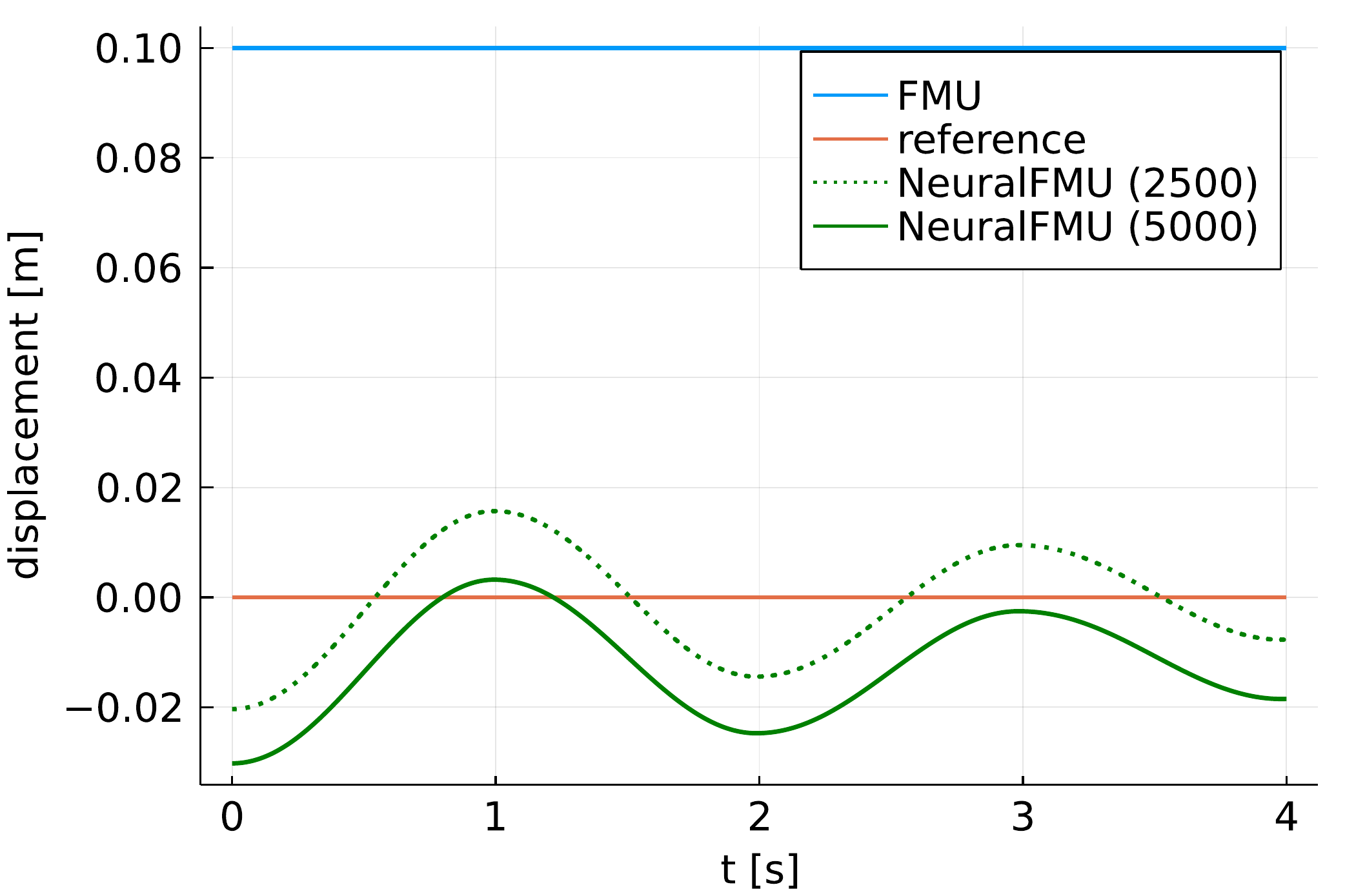}
	\caption{Comparison of the displacements of the \ac{FMU}, reference system and the NeuralFMU (top part \ac{NN}) after 2500 and 5000 training steps on testing data.}
	\label{fig:disp}
\end{figure}

The bottom part of the \ac{NN} learned the physical effect \emph{discontinuous friction} in a generalized way, because the net was trained based on the state derivatives instead of the states themselves. A comparison of the friction model of the reference system, the \ac{FMU} and the learned friction model, extracted from the bottom part \ac{NN} of the Neural\ac{FMU}, is shown in \autoref{fig:fric}. The learned friction model is a simplification of the reference friction model, because of the small net layout and a lack of data at the discontinuity near $v=0$. Finally, also the displacement modeling failure of the white-box model (\ac{FMU}) was canceled out by the small top \ac{NN} as can be seen in \autoref{fig:disp}.

\section{Conclusion}
The presented open source library \libfmi{} (\urlfmi) allows the easy and seamless integration of \ac{FMI}-models into the Julia programming language. \acp{FMU} can be loaded, parameterized and simulated using the abilities of the \ac{FMI}-standard. Optional functions like retrieving the partial derivatives or manipulating the \ac{FMU} state are available if supported by the \ac{FMU}. The library release version 0.1.4 is compatible with \ac{FMI} 2.0.x (the common version at the time of release), supporting upcoming standard updates like \ac{FMI} 3.0 is planned. The library currently supports \ac{ME}- as well as \ac{CS}-\acp{FMU}, running on Windows and Linux operation systems. Event-handling to simulate discontinuous \ac{ME}-\acp{FMU} is supported. 

The library extension \libfmiflux{} (\urlfmiflux{}) makes \acp{FMU} differentiable and opens the possibility to setup and train Neural\acp{FMU}, the structural combination of a \ac{FMU}, \ac{NN} and a numerical solver. Proper event-handling during back-propagation whilst training of Neural\acp{FMU} is under development, even if there were no problems during training with the discontinuous model from the paper example. A cumulative publication is planned, focusing on a real industrial use-case instead of a methodical presentation. 

Current and future work covers the implementation of a more general custom adjoint, meaning despite \libzygote{}, other \ac{AD}-frameworks will be supported. Further, we are working on different fall-backs if the optional function \mycode{fmi2GetDirectionalDerivatives} is not available. The finite differences approach for \ac{ME}-\acp{FMU} is already implemented, sampling via \mycode{fmi2GetState} and \mycode{fmi2SetState} for \ac{CS}-\acp{FMU} will be supported soon.

\acp{FMU} contain the model as compiled binary, therefore \ac{FMU} related computations must be performed on the CPU. On the other hand, deploying \acp{NN} on optimized hardware like GPUs often results in a better training performance. Currently, the training of the Neural\ac{FMU} is completely done on the CPU. A hybrid hardware training loop with the \ac{FMU} on the CPU and \ac{NN} on the GPU may lead to performance improvements for wider and deeper \ac{NN}-topologies.

An extension of the library to the \ac{CS}-standard \ac{SSP} \cite{SSP:2019}, including the necessary machine learning back-end, is near completion. This will allow the integration of complete \acp{CS} into a \ac{NN} topology and retrieve a \emph{NeuralSSP}.

Beside NeuralFMUs, \libfmiflux{} paves the way for other hybrid modeling techniques and new industrial use-cases by making \acp{FMU} differentiable in an \ac{AD}-framework. The authors are excited about any assistance they can get to extend the library repositories by new features and maintain them for the upcoming technology progress. Contributors are welcome.

\section*{Acknowledgments}
The authors like to thank \emph{Andreas Heuermann} for his hints and nice feedback during the development of the library. Further, we thank \emph{Florian Schläffer} for designing the beautiful library logos.
This work has been partly supported by the ITEA 3 cluster programme for the project UPSIM - Unleash Potentials in Simulation.

\printbibliography

\end{document}